\documentclass[10pt, a4paper]{article}
\usepackage{lrec2022} 
\usepackage{multibib}
\newcites{languageresource}{Language Resources}
\usepackage{graphicx}
\usepackage{tabularx}
\usepackage{soul}
\usepackage{titlesec}
\titleformat{\section}{\normalfont\large\bfseries\center}{\thesection.}{1em}{}
\titleformat{\subsection}{\normalfont\SmallTitleFont\bfseries\raggedright}{\thesubsection.}{1em}{}
\titleformat{\subsubsection}{\normalfont\normalsize\bfseries\raggedright}{\thesubsubsection.}{1em}{}
\renewcommand\thesection{\arabic{section}}
\renewcommand\thesubsection{\thesection.\arabic{subsection}}
\renewcommand\thesubsubsection{\thesubsection.\arabic{subsubsection}}

\usepackage{epstopdf}
\usepackage[utf8]{inputenc}

\usepackage{hyperref}
\usepackage{xstring}

\usepackage{color}

\usepackage{listings}
\usepackage{xcolor}
\usepackage{graphicx}
\usepackage{array}
\newcolumntype{L}[1]{>{\raggedright\let\newline\\\arraybackslash\hspace{0pt}}m{#1}}
\newcolumntype{C}[1]{>{\centering\let\newline\\\arraybackslash\hspace{0pt}}m{#1}}
\newcolumntype{R}[1]{>{\raggedleft\let\newline\\\arraybackslash\hspace{0pt}}m{#1}}
\newcolumntype{B}[1]{>{\raggedright\let\newline\\\arraybackslash\hspace{0pt}}p{#1}}
\newcolumntype{N}[1]{>{\centering\let\newline\\\arraybackslash\hspace{0pt}}p{#1}}
\newcolumntype{M}[1]{>{\raggedleft\let\newline\\\arraybackslash\hspace{0pt}}p{#1}}

\definecolor{codegreen}{rgb}{0,0.6,0}
\definecolor{codegray}{rgb}{0.5,0.5,0.5}
\definecolor{codepurple}{rgb}{0.58,0,0.82}
\definecolor{backcolour}{rgb}{0.95,0.95,0.92}

\lstdefinestyle{mystyle}{
    commentstyle=\color{codegreen},
    keywordstyle=\color{magenta},
    numberstyle=\tiny\color{codegray},
    stringstyle=\color{codepurple},
    basicstyle=\ttfamily\scriptsize\bfseries,
    breakatwhitespace=false,         
    breaklines=true,                 
    captionpos=b,                    
    keepspaces=true,                 
    numbers=left,                    
    numbersep=5pt,                  
    showspaces=false,                
    showstringspaces=false,
    showtabs=false,                  
    tabsize=2,
    xleftmargin=2em,
    frame=single,
    framexleftmargin=1.5em
}

\usepackage{float}
\newfloat{lstfloat}{t}{lop}
\floatname{lstfloat}{Listing}

\usepackage{textcomp}
\usepackage{microtype}
\usepackage{xspace}
\usepackage{multirow}
\usepackage{multicol}
\usepackage{makecell}
\usepackage{booktabs}
\usepackage{cleveref}
\crefformat{section}{\S#2#1#3} 
\crefformat{subsection}{\S#2#1#3}
\crefformat{subsubsection}{\S#2#1#3}

\usepackage{xcolor}
\newcommand{\as}[1]{[\texttt{#1}]} 
\newcommand{\grasp}{GrASP\xspace}

\title{\grasp{}: A Library for Extracting and Exploring\\ Human-Interpretable Textual Patterns}

\name{Piyawat Lertvittayakumjorn$^1$, Leshem Choshen$^2$, Eyal Shnarch$^2$, Francesca Toni$^1$} 

\address{$^1$Department of Computing, Imperial College London, United Kingdom\\
$^2$IBM Research\\
pl1515@imperial.ac.uk, 
\{leshem.choshen, eyals\}@il.ibm.com,
ft@imperial.ac.uk}

\abstract{
Data exploration is an important step of every data science and machine learning project, including those involving textual data. 
We provide a novel language tool, in the form of a publicly available Python library for extracting patterns from textual data.
The library integrates a first public implementation of the existing \grasp{} algorithm. It allows users to extract patterns using a number of general-purpose \emph{built-in} linguistic attributes (such as hypernyms, part-of-speech tags, and syntactic dependency tags), as envisaged for the original algorithm, as well as domain-specific \emph{custom} attributes which can be incorporated into the library by implementing two functions.  
The library is equipped with a web-based interface empowering human users to conveniently explore data via the extracted patterns, using complementary \emph{pattern-centric} and \emph{example-centric} views: the former includes a reading in natural language and statistics of each extracted pattern; the latter shows applications of each extracted pattern to training examples.
We demonstrate the usefulness of the library in classification 
(spam detection and argument mining), model analysis (machine translation), and artifact discovery in datasets (SNLI and 20Newsgroups). 
 \\ \newline   }

\begin{document}

\maketitleabstract

\section{Introduction}
\label{sec:intro}

\begin{table*}[t]
\small
\begin{tabular}{@{}cclc@{}}
    Use case & No. &  Sentences Matched & Pattern  \\ 
    \cmidrule[\heavyrulewidth](l){1-4}    

    \begin{tabular}{@{}c@{}} SMS\\ Spam\end{tabular} &
    \begin{tabular}{@{}c@{}} 1 \\2\\3 \end{tabular} &
     
    \begin{tabular}{@{}l@{}}  You are \textbf{awarded} \textbf{a} \textbf{SiPix} Digital Camera...\  \\ 
    ...to \textbf{WIN} \textbf{a} FREE \textbf{Bluetooth} Headset... \\ 
    ...for \textbf{Free} ! Call \textbf{The} \textbf{Mobile} Update... \end{tabular} &
     
     \begin{tabular}{@{}c@{}} [\as{SENTIMENT:pos}, \as{POS:det}, \as{POS:propn}] \\
     
     (A positive-sentiment word, closely followed by \\ a determiner, and then by a proper noun)
     \end{tabular}  \\ 
    \cmidrule[\lightrulewidth](l){1-4}
    
    \begin{tabular}{@{}c@{}} Topic-\\Dependent\\ Argument \\Mining \end{tabular} &   
    \begin{tabular}{@{}c@{}} 4 \\5\\6 \end{tabular} &

    \begin{tabular}{@{}l@{}} \textbf{Evidence} \textbf{suggests} \textbf{that} TOPIC is...   \\
    \textbf{Poll} in 2002 \textbf{found} \textbf{that} 58\% of... \\
    ...\textbf{data} \textbf{indicate} \textbf{that} TOPIC reduces...  \end{tabular} &
    
    \begin{tabular}{@{}c@{}} [\as{ARGUMENTATIVE}, \as{POS:verb}, \\ \as{LEMMA:that, POS:adp}] \\
    
    (An argumentative word, closely followed by a verb,\\ and then by the word ``that'' which is also a preposition)
    \end{tabular} \\
   
    \cmidrule[\lightrulewidth](l){1-4}
    
    \begin{tabular}{@{}c@{}} Machine \\ Translation \\ model analysis \end{tabular} &   
    \begin{tabular}{@{}c@{}} 7\\8\\9 \end{tabular} &

    \begin{tabular}{@{}l@{}} \textbf{choose} \textbf{Windows} \textbf{$>$} Channels \\
    \textbf{select} \textbf{Control} \textbf{$>$} Test Movie \\
    \textbf{select} Modify $>$ \textbf{Transform} \textbf{$>$} Skew \\
     \end{tabular} &
    
    \begin{tabular}{@{}c@{}} [\as{HYPERNYM:choose}, \as{DEP:dobj}, \as{LEMMA:>}] \\
    
    (A type of \textit{choose} (verb), closely followed by a word \\ with the dependency type \textit{dobj}, and then by \textit{$>$})
    \end{tabular} \\
   \cmidrule[\heavyrulewidth](l){1-4}
\end{tabular}

\caption{Examples of \grasp{} patterns capturing the common structure in a variety of surface forms -- the sentences matched.
Matched words are in bold. A description of each pattern is provided below it, in parentheses.}
\label{tab:pattern_exmples}
\end{table*}

Pattern-based analysis is a promising approach for exploring textual data in many tasks, such as authorship classification \cite{houvardas2006n}, relation extraction \cite{hearst-1992,peng2014generalizable} and argument mining \cite{madnani-etal-2012-identifying}.
Patterns uncover prominent characteristics of the data which lead to interesting insights, helping humans proceed to the next steps (e.g., data pre-processing, setting model hyperparameters).
In this paper, we implement and publicly release a Python library named \grasp{} after a supervised algorithm which learns expressive patterns 
(in the form of sequences of conjunctions of attributes)
characterizing
linguistic phenomena \cite{shnarch-etal-2017-grasp}. Apart from the usefulness of the web resource, this is the first public implementation of \grasp{}.
To illustrate the richness of GrASP patterns, examples 1--3 in Table \ref{tab:pattern_exmples} are all SMS spam messages from the dataset by \newcite{almeida2011contributions}. While there is little word overlap between them, their commonality is apparent, even if hard to name. 
The \grasp{} algorithm can reveal an underlying structure which generalizes these three realizations of spams within a single pattern: a positive-sentiment word (represented by \texttt{SENTIMENT:pos}), closely followed by a determiner (represented by \texttt{POS:det}, where \texttt{POS} stands for `part of speech'), and then by a proper noun (represented by \texttt{POS:propn}).

The input for the algorithm 
amounts to 
two sets of texts: in one (the positive set) the target phenomenon appears in all examples (e.g., all are spam messages), 
and in the other (the negative set) it does not. \grasp{} looks for commonalities prominent within the texts of one set but not  
shared across the
sets. To be able to recognize common aspects of texts, beyond their surface form realizations, all input tokens are augmented with a variety of linguistic attributes such as part-of-speech tags, named entity information, or pertinence to a lexicon (e.g., of sentiment words). Attributes are selected to maximize a score (by default their information gain about the label). Then, they are combined by a greedy algorithm to generate patterns which are most indicative either to the positive or the negative set. 

Since the patterns are a combination of readable attributes, they are human interpretable. So, they can be used to provide insights about the data and contribute to explainability.
For instance, example 1 in Table \ref{tab:pattern_exmples} is identified as a spam message because it contains the phrase \textit{awarded a SiPix}. 
Patterns can also be used for classification (e.g., if the pattern mentioned above is matched in a message, we can classify it as spam).

The contribution of this paper is threefold:
\begin{itemize}
    \item We release \grasp{}, a Python library for extracting interpretable patterns from text.
    Although the algorithm of GrASP was initially proposed in 2017, we are the first to make its implementation publicly available.\footnote{The library is available at \url{https://github.com/plkumjorn/GrASP}}
    The library provides additional extensions, allowing users to customize the algorithm to their specific needs (see \cref{sec:library}).
    
    \item We provide a novel web-based GUI which displays
    \grasp{} patterns and their matches in a dataset. This exploration tool allows users to conveniently explore the patterns and the data with both \emph{pattern-centric} views and \emph{example-centric} views and gain insights (see \cref{sec:webtool}).
    
    \item We exemplify the usefulness of the provided library
    in three use cases: (i) classification (see \cref{sec:classification}), (ii) model analysis (see \cref{sec:model_analysis}), and (iii) dataset artifacts unveiling (see \cref{sec:artifacts}).
    These use cases cover several applications including spam classification, argument mining, topic classification, natural language inference, and machine translation.

\end{itemize}

\section{\grasp Library for Pattern Extraction} \label{sec:library}

\begin{lstfloat}
\begin{lstlisting}[language=Python, caption=Basic usage of \grasp{}, label={lst:usage}]
from grasp import GrASP
# Step 1: Create the GrASP model
grasp_model = GrASP(num_patterns = 200, gaps_allowed = 2, alphabet_size = 200, include_standard = ['TEXT', 'POS', 'NER',   'SENTIMENT'])
# Step 2: Fit it to the training data
grasp_model.fit_transform(pos_exs, neg_exs)
# Step 3: Export the results 
grasp_model.to_csv('results.csv')
grasp_model.to_json('results.json')
\end{lstlisting}
\end{lstfloat}

Listing~\ref{lst:usage} shows the basic usage of \grasp{} for extracting patterns from training data, i.e., two lists of texts containing positive examples (\texttt{pos\_exs)} and negative examples (\texttt{neg\_exs}).
In the first step, the user specifies hyperparameters such as the desired number of patterns and the set of linguistic attributes for augmenting input tokens.
The following general-purpose attributes are built-in: the token text itself (in lower case), its lemma, wordnet hypernyms, 
part-of-speech tags, named-entity, syntactic dependency, and sentiment tags.\footnote{We use \href{https://spacy.io/}{spacy} for tokenization and tagging, \href{https://www.nltk.org/}{nltk} and lesk \cite{lesk1986automatic} for word sense disambiguation and finding hypernyms, and a sentiment lexicon by \newcite{hu2004mining}.
For hypernyms, we consider only three levels above the token of interest in order to avoid terms that are too abstract to comprehend (e.g., \textit{psychological feature}, \textit{group action}, \textit{entity}).

}  

After that, the user trains the \grasp{} model and exports the results to a csv file or a json file.
Both file types summarize all the extracted patterns and their relevant statistics, while the json file also contains the configuration, the alphabet, and the training examples annotated with the patterns matched.

Note that in our version of \grasp{}, we add several parameters which are not found in the original algorithm. 
While the original \grasp{} uses information gain as the criterion to rank and select patterns, \newcite{shnarch-etal-2020-unsupervised} preferred $F_{\beta}$ (where $\beta \leq 1$) as a criterion, emphasizing precision over recall. Hence, we allow the users to implement their own criteria, tailored to their use cases. 
Other new parameters for customizing \grasp{} include the number of gaps allowed between the matched tokens (which overrides the window size parameter) and the minimum coverage for a pattern to be selected. 

\subsection{Extending GrASP Capabilities} \label{ssec:new_features}
Besides the basic usage, we provide the following new useful features.

\paragraph{Translating patterns into natural language explanations.}
It could be difficult for lay users to read and understand linguistic attributes in patterns. Therefore, using templates, we provide a function \texttt{pattern2text} for translating a pattern (e.g., [\as{HYPERNYM:communication.n.02}, \as{POS:NUM}] with a window size of 4) into an English explanation (e.g., ``A type of communication (n), closely followed by a number''). 
This feature removes the barrier to an audience without linguistic knowledge to comprehend the resulting patterns.
We believe that it is very important in providing explainability for all, not just for NLP or machine learning experts.

\paragraph{Using custom attributes.}
In addition to the built-in attributes, users may require custom attributes, to better fit the patterns to their domain.
For instance, given biomedical texts, attributes indicating whether a token refers to a kind of biomedical substance (e.g., protein, drug, enzyme) could be useful for relation extraction from texts.
Our library allows users to create custom attributes by using the \texttt{CustomAttribute} class and implementing two functions. One extracts the attributes from an input text, and the other explains the attributes for the translation feature. 
We provide an example of custom attributes in the README file of our library's GitHub repository.

\paragraph{Removing redundant patterns.} 
We provide a function to remove specific patterns 
subsumed by more general ones. For instance, all [\as{TEXT:.}, \as{POS:NUM}] matches are also matches of [\as{POS:NUM}]. 
Hence, this function removes the former (specific) pattern.
A flag further allows to condition removal on the fact that the score of the removed pattern must be lower than the score of the remaining one.

\paragraph{Vectorizing texts using patterns.}
Given an input text $t$ and $n$ \grasp{} patterns, we provide a function to create a ternary vector of length $n$ indicating which of the patterns are matched in $t$ and whether they are positive or negative. This vector can be used as features of $t$ for downstream tasks.

\section{The Web-Based Exploration Tool} \label{sec:webtool}
\begin{figure*}[t] 
    \centering
    \includegraphics[trim={0 0.2cm 0 0},clip, width=\linewidth]{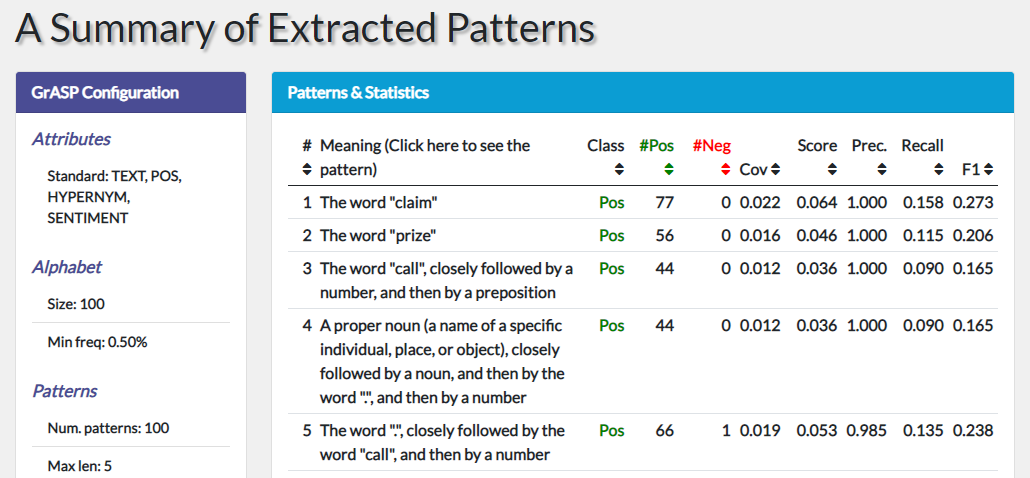}
    \caption{A part of the pattern-centric level 1 report listing the patterns and their metrics with respect to the training data, the configuration of \grasp{}, and the training set statistics.} \label{fig:view1}
\end{figure*}

\begin{figure*}[t] 
    \centering
    \includegraphics[width=\linewidth]{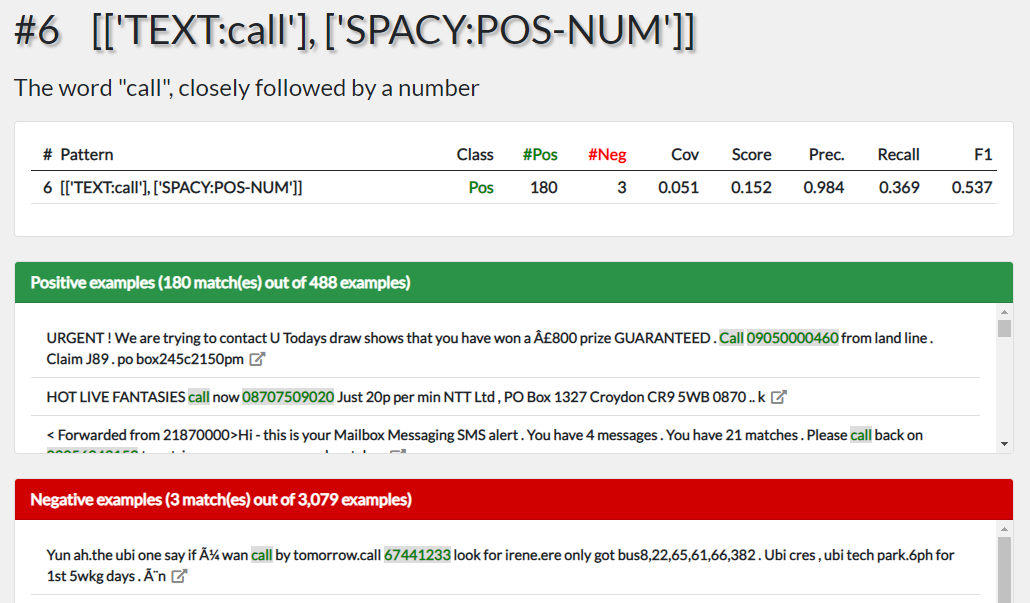}
    \caption{A part of the pattern-centric level 2 report listing positive and negative matches by a specific pattern.} \label{fig:view2}
\end{figure*}

\begin{figure*}[t] 
    \centering
    \includegraphics[width=\linewidth]{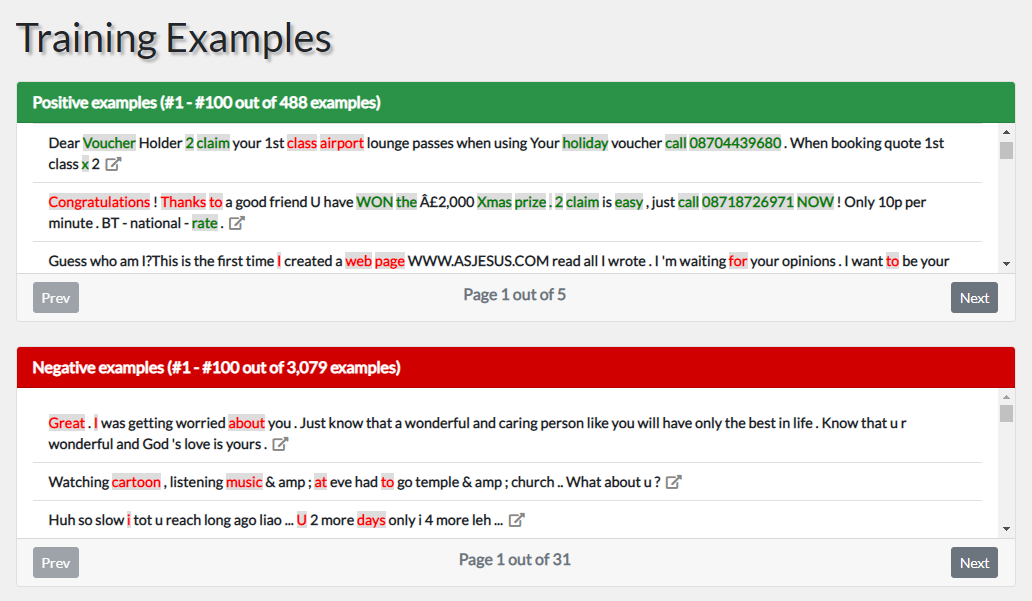}
    \caption{The example-centric level 1 report listing all positive and negative examples in the training data. Words matched by at least one pattern are highlighted.} \label{fig:view3}
\end{figure*}

\begin{figure*}[t] 
    \centering
    \includegraphics[trim={0 2.2cm 0 0},clip,width=\linewidth]{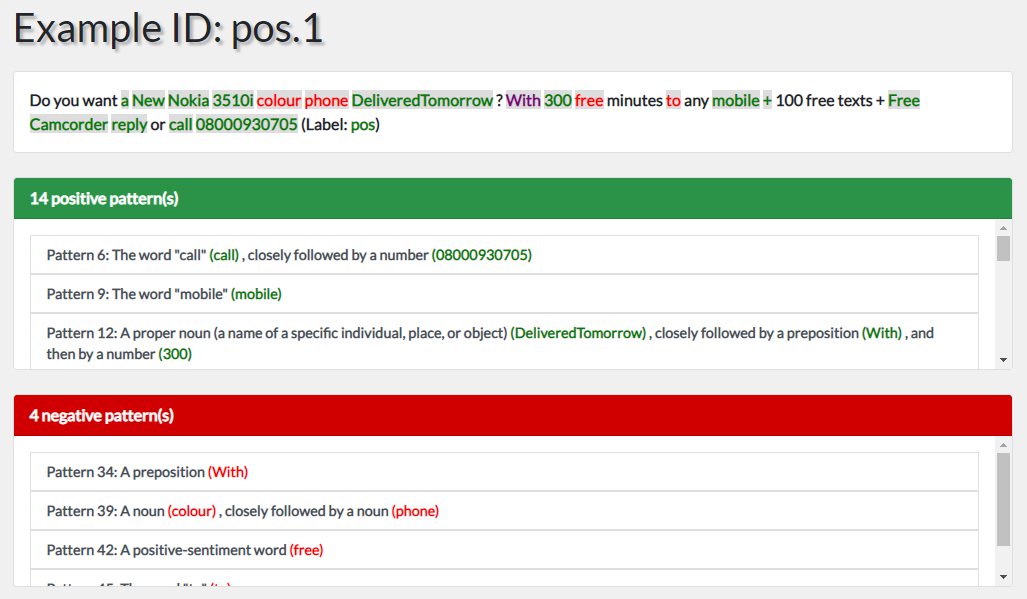}
    \caption{A part of the example-centric level 2 report showing positive and negative patterns matching an example.} \label{fig:view4}
\end{figure*}

We provide a web-based exploration tool implemented using Flask.\footnote{\url{flask.palletsprojects.com/en/1.1.x/}} Taking as input the json file exported by the \grasp{} library, this tool provides four types of reports: two are pattern-centric and the other two are example-centric. 
Figures~\ref{fig:view1}--\ref{fig:view4} show (parts of) these reports for the spam classification example.
Additionally, all reports for all the use cases mentioned in this paper are available online for further examination.\footnote{\url{https://plkumjorn.github.io/GrASP}}

\paragraph{
I. Pattern-centric level 1.} 
(Figure \ref{fig:view1}) This report lists all the patterns together with the configuration of \grasp{} used for extracting them and the statistics about the training data. 
For each pattern (row), its statistics on the training data are reported, including the number of positive and negative examples matched, coverage, the metric score, precision, recall, and F1. 
Users can click a header to sort the table based on the column values.
Moreover, users can click the header of the pattern column to translate the patterns into their natural language meanings (as shown in Figure~\ref{fig:view1}).
Once users click a pattern in the table, they will be redirected to that pattern's level 2 report.

\paragraph{
II. Pattern-centric level 2.}
(Figure \ref{fig:view2}) This report can be generated for each pattern individually. In addition to the pattern's statistics, it shows the positive and 
negative examples matched by this pattern, with the corresponding tokens highlighted.
Users can click the link icon at the end of each example to go to its example-centric level 2 report (see below).

\paragraph{
III. Example-centric level 1.}
(Figure \ref{fig:view3}) This report lists all positive and negative training examples (with pagination). For each example, words matched by only positive pattern(s) 
and only negative pattern(s) are highlighted in green and red, respectively, while words matched by both types of patterns are in purple.
By clicking a highlighted word, users can see the list of patterns matching this word. Clicking the link icon of an example directs to its example-centric level 2 report.

\paragraph{
IV. Example-centric level 2.}
(Figure \ref{fig:view4}) This report can be generated for each training example, showing all the positive and 
negative patterns matching it. Clicking a pattern links to its pattern-centric level 2 report.

Overall, these reports provide a convenient way for quickly exploring the data via the lens of the common patterns found in it. 
Also, we can use this exploration tool to display results from other pattern extraction algorithms, in addition to \grasp, provided that their results are organized into the specific json format required by this tool (as described in the README file of our \grasp GitHub repository).
Next, we will show how insights can be gained from this exploration tool with outputs from the \grasp algorithm.

\section{Use Case: Classification} \label{sec:classification}
Apart from the spam classification example discussed throughout the paper, we present another classification use case of our tools
targeting the corpus of topic-dependent argument mining \cite{shnarch2018will}.
Generally, argument mining aims to identify and extract the structure of inference and reasoning expressed as arguments in texts \cite{lawrence2020argument}.
The dataset used here in particular focuses on identifying, in text, evidence for claims. 
It includes 4,065 training and 1,720 test pairs of topics and sentences, each labeled with whether the sentence is a piece of valid evidence for the topic.

We augment each token with the built-in attributes 
as well as a custom binary attribute indicating whether a token was found in a lexicon of argumentative words. We use information gain to select patterns, set the number of patterns to 100, and allow up to 2 gaps in each pattern. 
Next, we train and examine the patterns with the pattern-centric reports. 

We find that the most indicative pattern is the word \textit{that} in its \textit{preposition} meaning (as opposed to other meanings of that word, e.g., a determiner, as in this sentence). This is a very nice outcome since \grasp{} automatically revealed what experts have found -- ``that'' is an indicator of argumentative content and can be
used as an effective source of weak supervision \cite{levy2017unsupervised}. Other informative patterns are about studies: presenting them (e.g., [\as{POS:NUM},\as{LEMMA:study}] which matches phrases like \textit{\textbf{one study}}, \textit{\textbf{Two} major \textbf{studies}}, \textit{A \textbf{1999} meta \textbf{study}}, and \textit{a \textbf{25} year longitudinal \textbf{study}}), declaring their findings (e.g., examples 4--6 in Table \ref{tab:pattern_exmples}), and reporting their experimental results (e.g., [\as{POS:ADP},\as{POS:VERB},\as{HYPERNYM:risk}] which covers expressions such as \textit{\textbf{in} \textbf{reducing} the \textbf{risk}}, \textit{more \textbf{than} \textbf{tripled} the \textbf{risk}}, and \textit{\textbf{as} \textbf{controlling} the \textbf{risks}}). These patterns, too, have rediscovered what is known in the domain 
-- texts describing studies are a good source for evidence \cite{rinott-etal-2015-show,Ein-Dor-2020}.

The exploration tool enables having a human in the loop by making it easy to compare patterns (e.g., by their recall-precision trade off), examine 
their realizations in the positive and negative sets, check pattern combination (e.g., click on a purple word in the example-centric report to see all patterns that matched it), and evaluate the contribution of adding custom attributes (such as the lexicon of argumentative words). 
After that, users can improve on their own the list of patterns by filtering, correcting, and merging patterns. 
The selected patterns can then contribute to the downstream pipeline (e.g., by using the most precise/high-coverage patterns to improve precision/recall). 
Alternatively, the patterns can be used to obtain weak supervision, combined with labeled data,
to improve performance as in \cite{shnarch2018will}. Finally, similar to \cite{sen-etal-2020-learning}, the human-readable patterns 
matching an example can provide explainability for the model decision for it.

\section{Use Case: Model Analysis} \label{sec:model_analysis}
In this use case, we show how \grasp{} library can assist the process of analyzing an NLP model, and specifically, finding its strength and weaknesses.
While for some tasks, such as Parsing \cite{nilsson2008malteval} or Grammatical Error Correction \cite{choshen2020classifying}, exploring model's decisions naturally supports its analysis (e.g., which types of grammatical errors the model does not cover well), in many tasks this is not the case. We take Machine Translation (MT) as an example where characterizing model performance poses a challenge  \cite{stanovsky2019evaluating,Renduchintala2021InvestigatingFO}. We demonstrate how \grasp{} library can be used to identify both challenging and easy inputs for a given MT model. An example of the practical usage of such analysis is found in translation systems which combine automatic MT and human translation. If we can identify that some types of inputs are translated well by the model, then we can refer them to the MT and manually translate only the rest. This will lower the system costs and aid in focusing human effort in the difficult translation cases and revision \cite{hutchins2001machine,taravella2013acknowledging}. 

We use the WMT19 English-German Quality Estimation data \cite{fonseca2019WMT} which  consists of the automatic translation of 13K sentences by a specific model. Each translated sentence was judged by humans and was assigned with a translation quality score. 
We sort all sentences by this quality score and take the top 25\% (inputs on which the model performed best) as the positive set for \grasp{}, and the bottom 25\% as the negative set. We train \grasp{} with the default hyperparameters and output 100 patterns.\footnote{The full list of GrASP hyperparameters for each use case in this paper can be found in the pattern-centric level 1 report of the use case on our live demo website.}

First, we focus on indicative patterns in the positive set, i.e., the easiest inputs for the model. There are 60 highly precise patterns (precision $\geq$ 80\%), each with recall ranging from 5\% up to 35\%. Among those, we find various patterns which indicate that instruction texts are easy to translate. 
These patterns include the sign \textit{$>$}, the lemmas \textit{click} and \textit{select}, the \textit{hypernym of choose} (which matches both \textit{choose} and \textit{select}).
For an example of such pattern and its matches, see lines 7--9 in Table \ref{tab:pattern_exmples}. 
Furthermore, by examining these instruction patterns, we did notice that the dataset is composed of two specific domains (Reviews and IT), 
unlike several other machine translation datasets that consist of
general domain data, such as news and Wikipedia \cite{barrault2020findings,specia2020WMT2}. 
This accidentally exemplifies the power of the exploration tool to teach us about the data we use.

Revealing which inputs are easy for the translation model is important, but so is the opposite. Various studies have analyzed what challenges translation models to aid their improvement \cite{macketanz2018fine,choshen2019automatically}. Hence, we observe the negative-set patterns, i.e., the most challenging inputs.
\grasp{} finds much less such patterns, suggesting that the challenging inputs are more diverse and harder to classify. Still, 16 patterns with precision larger than 60\% are found. 

Of specific interest to us are patterns that match inputs that are known in the literature to be challenging. We report two such examples. \newcite{Gralinski2019GEvalTF} highlight the lemma \textit{be} as an especially hard case. Indeed, \grasp{} dedicates a special pattern for this lemma. Another known difficult input is that of structural ambiguity \cite{avramidis2019linguistic}. For example, \textit{new blog entry}, where the translation model needs to decide which is new, the \textit{blog entry} or the \textit{blog} itself. \grasp{} finds the pattern [\as{POS:ADJ}, \as{POS:NOUN}, \as{POS:NOUN}] which captures these cases.

Overall, this use case shows that the library is beneficial for analyzing models by extracting known hurdles and identifying easy cases. Other uses for analysis may include different tasks or comparing models to each other.

\section{Use Case: Dataset Artifacts}
\label{sec:artifacts}
Artifacts in text classification datasets are tokens or phrases which are irrelevant, but frequently appear in examples of some of the classes. 
Hence, the trained models could rely on spurious correlations caused by these artifacts to make predictions.
This is usually undesirable since it prevents the models from generalizing well out of the training distribution.
In this use case, we leverage \grasp to help humans detect such artifacts in training datasets.

\paragraph{SNLI.} Natural language inference is a task aiming to predict whether
a hypothesis statement is true (entailment), false (contradiction), or undetermined (neither) given a premise statement.
Using the pointwise mutual information (PMI), \newcite{gururangan-etal-2018-annotation} discovered annotation artifacts in many hypotheses of the Stanford Natural Language Inference (SNLI) dataset \cite{bowman-etal-2015-large}.
These artifacts make the fasttext classifier \cite{joulin-etal-2017-bag} classify examples correctly for 67.0\% without even looking at the premises (while the majority class baseline is only 34.3\%).

We applied \grasp (with the number of patterns and the alphabet size of 200, using information gain as the selection criteria) to a subset of hypotheses in SNLI, i.e., 10K sentences from each of the entailment (positive) and contradiction (negative) classes.
Then we kept only patterns with precision $\geq$ 75\%.
Among the final set of patterns, we found artifacts reported by \newcite{gururangan-etal-2018-annotation} yet with additional contexts and beyond. 
While PMI found that \textit{outside} and \textit{outdoors} are important words for the entailment class, \grasp discovered the pattern [\as{POS:NOUN}, \as{LEMMA:be}, \as{HYPERNYM:outside}] with the precision of 97\% for the entailment class.
This pattern covers both \textit{outside} and \textit{outdoors} (thanks to the \texttt{HYPERNYM} attribute) and further reveals their semantic role in the sentences.
Another similar pattern, this time for the contradiction class, is [\as{HYPERNYM:person}, \as{HYPERNYM:be},  \as{POS:VERB}, \as{LEMMA:in,POS:ADP}] (matching, e.g., \textit{man/women/girl/hiker/Firefighters is/are lying/bored/swimming/reading in} as shown in the pattern-centric level 2 report). It shows that many contradicting hypotheses were crafted by indicating the actions (verbs) a human took and their locations (following the word \textit{in}).
Besides, the pattern [\as{HYPERNYM:nutriment,POS:NOUN}], grouping \textit{lunch}, \textit{pizza}, \textit{dessert}, \textit{picnic}, etc. into one category, helps us quickly grasp the artifact theme.
Interestingly, we also found from the pattern [\as{SENTIMENT:neg,POS:VERB}] that verbs with negative sentiments such as \textit{attack}, \textit{stole}, \textit{lost} and \textit{fell} fairly correlate with the contradiction class (with 75.7\% precision).
All in all, a strength of \grasp is the ability to group words based on their semantics or functions. Together with the ordered representation of patterns, it provides better understanding of the artifacts.

\paragraph{20Newsgroups} 
\cite{Lang95}
is another dataset containing artifacts. Here, we aim to not only identify the artifacts but also partially remove them to make the classifier generalize better.
Particularly, we focus on distinguishing the Christianity class from the two related classes, Atheism and Religion (miscellaneous).
The goal is to improve generalization of a model trained on the 20Newsgroups dataset and tested on the Religion dataset \cite{ribeiro2016lime}.

To detect the artifacts, we applied \grasp using
Christianity as the positive class and the other two classes as the negative class.
Generally, if the positive and the negative examples are not balanced, \grasp's precision might be biased towards the majority class.
Hence, we downsampled the larger class (non-Christianity) to reach the same size as the smaller class. 
We then kept only patterns with at least 75\% precision, resulting in 133 patterns.

After that, as the patterns are human-interpretable,
we manually checked them using the web exploration tool and flagged patterns 
that are semantically irrelevant to the classification task (i.e., likely to be artifacts).
The pattern-centric view level 1 shows all the patterns to be annotated. 
When being unsure what a pattern covers, we used the pattern-centric view level 2 to see matched examples before making a decision. 
For example, one can see that the pattern [\as{HYPERNYM:sacrament}] covers \textit{liturgy}, \textit{baptism}, \textit{confession}, and \textit{Eucharist}, and accepting this pattern helps us annotate words or phrases in bulk.
Overall, 40 out of 133 patterns were marked irrelevant.
These include patterns for irrelevant names and surnames  (e.g., [\as{LEMMA:keith}], [\as{LEMMA:mozumder}]), 
for generic words (e.g., [\as{LEMMA:cheers}], [\as{LEMMA:recent}]), 
and for punctuation (e.g., [\as{LEMMA:>}, \as{LEMMA:\#}]).

With the artifact patterns identified,
there could be several ways to address this problem and enhance model generalizability. 
Since the main objective of this paper is to demonstrate the applications of \grasp, we opted for simplicity and removed all the training examples that matched at least one artifact pattern. 
Then we used this filtered dataset to train a 1D Convolutional Neural Network \cite{kim-2014-convolutional} with 300-dimensional GloVe embeddings \cite{pennington-etal-2014-glove}.
We compared this model with another model trained on a sample of the original dataset matching the filtered dataset in the number of positive and negative examples (to disregard the effect of the dataset size). 
For each model, we ran the train and test process five times and averaged the results, reported in Table~\ref{tab:artifact_results}.

We can see that, both models perform almost equally well on the in-domain test set (from 20Newsgroups).
However, the model trained on the filtered dataset performs better on the out-of-distribution test set (about 5\% and 6\% absolute difference for the accuracy and macro F1, respectively). 
This demonstrates that \grasp can help humans identify artifacts in training data which is a prerequisite step for building a more generalizable classification model.

\begin{table}[t!] 
    \centering
    \begin{tabular}{|L{0.20\linewidth}|C{0.29\linewidth}|C{0.29\linewidth}|}
    \hline
    Setting & Accuracy & Macro F1\\\hline
    \multicolumn{3}{|c|}{Test dataset: 20Newsgroups} \\\hline
    Sampled & \textbf{0.810 $\pm$ 0.03} & \textbf{0.818 $\pm$ 0.03}	\\
    Filtered & 0.800 $\pm$ 0.01 & 0.804 $\pm$ 0.01	\\\hline
    \multicolumn{3}{|c|}{Test dataset: Religion (Out-of-distribution)} \\\hline
    Sampled & 0.674 $\pm$ 0.03 & 0.682 $\pm$ 0.03\\
    Filtered & \textbf{0.725 $\pm$ 0.06} & \textbf{0.739 $\pm$ 0.05}\\
    \hline
    \end{tabular}
    \caption{The results of the data artifacts (20Newsgroups) use case (average of five repetitions $\pm$ SD).} \label{tab:artifact_results}
\end{table}

\section{Related Work} \label{sec:rel}
\paragraph{Textual Data Exploration Tools.} 
While there is need for data exploration and interpretable patterns, not many tools exist. 
Most of the data exploration tools for text analyze the given data only at the word level \cite{heimerl2014word,zainol2018visualurtext} or the n-gram level \cite{davies2012word,benoit2018quanteda}.
Even though there exist tools extracting complicated patterns (like regular expressions) \cite{bartoli2016inference,locascio-etal-2016-neural}, the extracted patterns are not generalizable, i.e., having limited capability to match tokens based on their linguistic attributes.
To the best of our knowledge, apart from our work, there is no other publicly available tool to extract and display recurring generalizable patterns from text.
Somewhat related to ours, there are pattern-based search engines over data, even ones with web interfaces \cite{Shlain2020SyntacticSB,Resnik2005TheLS}. However, such resources require an expert to come up with the patterns herself in order to query data. Unlike our library, those patterns are not learned by the resources as a helping mechanism for the data exploration.
In other cases, patterns are not supplied by an expert. They are however, based on preexisting rules to extract specific information from the data, rather than to understand the data and what patterns occur in it \cite{ValenzuelaEscarcega2020OdinsonAF,ValenzuelaEscarcega2015ADR}.

Despite the fact that there are other algorithms except \grasp{} to learn patterns, such as \cite{sen-etal-2020-learning}, we could not find their tools or implementations publicly released.  It is noteworthy that the outputs of such algorithms could be formatted and plugged into our web-based exploration tool as well.

\paragraph{Debugging Tools for NLP.}
Although \grasp was devised for textual pattern extraction and exploration in general, it can be applied to debug NLP models and datasets as shown in sections \ref{sec:model_analysis} and \ref{sec:artifacts}, respectively.
Since debugging is a crucial and widely studied topic \cite{lertvittayakumjorn2021explanation}, there are several tools available to help humans debug NLP models and datasets.
Some of them are similar to our use cases.
For example, \newcite{Gralinski2019GEvalTF} propose GEval, which highlights features in inputs, outputs, or expected outputs that often make the model struggle.
This helps users notice the model weaknesses. However, GEval does not consider the order of features in text, so it requires users' effort to form patterns from feature combinations. 
\newcite{wu-etal-2019-errudite} propose Errudite, which extracts attributes from input instances and allows expert users to write a rule with the attributes so as to query and group model errors, enabling further model behavior analysis.
Concerning dataset artifact discovery, existing works usually focus on word-level artifacts  \cite{gururangan-etal-2018-annotation,gardner-etal-2021-competency}. Meanwhile, when it comes to more complex artifacts, humans are required to notice or hypothesize those artifacts with or without supporting tools \cite{mccoy-etal-2019-right,han-etal-2020-explaining}.
\grasp, in contrast, learns and provides patterns which might be artifacts to humans, so they only need to check those patterns with the aid of our web-based exploration tool, leading to more efficiency in the artifact discovery process for complicated patterns. 

\section{Conclusion and Outlook}
\label{sec:conclusion}
We presented \grasp{} -- a Python library for extracting patterns from textual data together with the novel web-based tool for exploring the resulting patterns and the input examples. Also, we demonstrated the usefulness of both via three use cases in NLP research (classification, machine translation model analysis, and data artifacts identification).  
Apart from these, \grasp{} has potential to be applied to several other problems such as 
explainable text classification (where patterns provide explanations) \cite{efstathiadis2022explainable,ribeiro2018anchors},
human-AI model co-creation (with patterns enabling communication) \cite{yang2019study}, and
interpreting deep learning models 
(e.g., by studying patterns that the models' neurons capture as reflected by their activation scores) \cite{lertvittayakumjorn-etal-2020-find,albini2020dax}.
Furthermore, the \grasp{} algorithm and the provided exploration tool are language-agnostic, so, as future work, it would be interesting to apply \grasp{} to a language besides English. To do so,  users need to create custom attributes that augment input tokens with linguistic features of the target language.
More generally, \grasp{} is applicable to other types of sequence data beyond text as long as we can extract suitable attributes for them \cite{Agrawal:95}.
Lastly, it is possible and interesting to extend our library with the functionalities of  \grasp{}$^{lite}$ \cite{shnarch-etal-2020-unsupervised} –- 
extracting patterns in an unsupervised way when we have a single list of texts rather than two lists (i.e., positive and negative classes) of texts.

\section{Acknowledgements}
The first author wishes to thank the support from Anandamahidol Foundation, Thailand.
The last author was partially supported by funding from the European Research Council (ERC) under the European Union’s Horizon 2020 research and innovation programme (grant agreement No. 101020934, ADIX, Argumentation-based Deep Interactive eXplanations).


\section{Bibliographical References}\label{reference}

\bibliographystyle{lrec2022-bib}
\bibliography{anthology,bib}


\end{document}